\documentclass[letterpaper]{article}
\usepackage{aaai2026}
\usepackage{times}
\usepackage{helvet}
\usepackage{courier}
\usepackage[hyphens]{url}
\usepackage{graphicx}
\usepackage{natbib}
\usepackage{caption}
\usepackage{algorithm}
\usepackage{newfloat}
\usepackage{listings}
\usepackage{amsmath}

\usepackage{xcolor}

\usepackage{algpseudocode}

\usepackage{amsmath,amsfonts,amssymb}
\usepackage{mathtools}  
\usepackage{bm}         
\usepackage{booktabs}
\usepackage{amsthm}
\theoremstyle{definition}

\theoremstyle{plain}

\theoremstyle{remark}

\DeclareCaptionStyle{ruled}{labelfont=normalfont,labelsep=colon,strut=off}
\floatstyle{ruled}
\newfloat{listing}{tb}{lst}{}
\floatname{listing}{Listing}
\title{ Automatic Failure Attribution and Critical Step Prediction Method for Multi-Agent Systems Based on Causal Inference}
\author{
    \textbf{Guoqing Ma}\textsuperscript{1},
    \textbf{Jia Zhu}\textsuperscript{2},
    \textbf{Hanghui Guo}\textsuperscript{1},
    \textbf{Weijie Shi}\textsuperscript{3},
    \textbf{Jiawei Shen}\textsuperscript{1},
    \textbf{Jingjiang Liu}\textsuperscript{1},
    \textbf{Yidan Liang}\textsuperscript{1}
}
\affiliations{
    \textsuperscript{1}School of Computer Science and Technology, Zhejiang Normal University, Zhejiang, China\\
    \textsuperscript{2}Zhejiang Key Laboratory of Intelligent Education Technology and Application, Zhejiang Normal University, Zhejiang, China\\
    \textsuperscript{3}Department of Computer Science and Engineering, Hong Kong University of Science and Technology, Hong Kong, China
}

\begin{document}
\maketitle

\begin{abstract}
Multi-agent systems (MAS) are critical for automating complex tasks, yet their practical deployment is severely hampered by the challenge of failure attribution. Current diagnostic tools, which rely on statistical correlations, are fundamentally inadequate; on challenging benchmarks like Who\&When, state-of-the-art methods achieve less than 15\% accuracy in locating the root-cause step of a failure. To address this critical gap, we introduce the first failure attribution framework for MAS grounded in multi-granularity causal inference. Our approach makes two key technical contributions: (1) a performance causal inversion principle, which correctly models performance dependencies by reversing the data flow in execution logs, combined with Shapley values to accurately assign agent-level blame; (2) a novel causal discovery algorithm, CDC-MAS, that robustly identifies critical failure steps by tackling the non-stationary nature of MAS interaction data. The framework's attribution results directly fuel an automated optimization loop, generating targeted suggestions whose efficacy is validated via counterfactual simulations. Evaluations on the Who\&When and TRAIL benchmarks demonstrate a significant leap in performance. Our method achieves up to 36.2\% step-level accuracy. Crucially, the generated optimizations boost overall task success rates by an average of 22.4\%. This work provides a principled and effective solution for debugging complex agent interactions, paving the way for more reliable and interpretable multi-agent systems.
\end{abstract}


\begin{figure}[t]
\centering
\includegraphics[width=0.9\columnwidth]{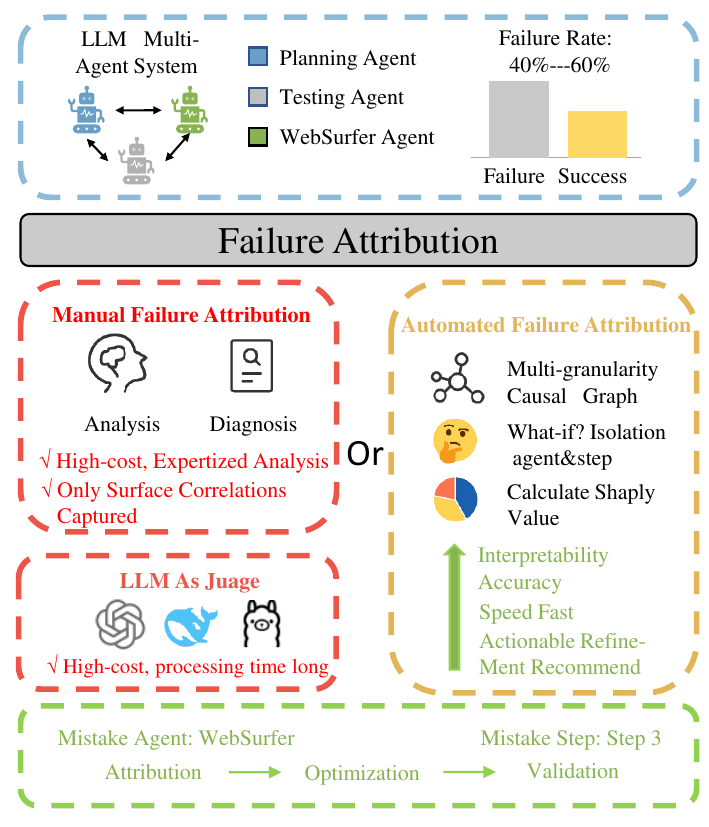} 
\caption{Challenges in Fault Attribution for Multi-Agent Systems (MAS) and Our Causal Framework. This figure contrasts the current diagnostic methods (left) with our proposed automated causal attribution process (right), which significantly enhances the accuracy, speed, and operability of attribution.}
\label{fig1}
\end{figure}
\section{Introduction}

Large Language Model (LLM)-powered Multi-Agent Systems (MAS) \cite{wang2024survey,xi2025rise,park2023generative} have demonstrated tremendous potential in automating complex workflows—from collaborative coding to scientific discovery—through task decomposition, inter-agent negotiation, and dynamic tool utilization~\cite{wang2023survey,liu2023agentbench,hong2023metagpt,qian2023chatdev}. However, real-world deployments reveal a stark reality: MAS not only exhibit failure rates as high as 40--60\%, but their debugging and optimization costs account for up to 70\% of total maintenance efforts~\cite{zhang2025agent,deshpande2025trail}. More critically, in many standard benchmarks, MAS performance fails to surpass single-agent baselines~\cite{pan2025multiagent,gao2025single}, highlighting fundamental gaps in reliability and scalability. These failures are not merely simple programming errors, but rather systemic, emergent issues arising from complex inter-agent interactions, coordination breakdowns, and error propagation~\cite{kalech2022model,herrera2020multi}.

Currently, the core bottleneck hindering MAS reliability lies in the absence of \textbf{automated failure attribution} \cite{zhuge2024agent}: when a MAS execution fails, how can we systematically identify which agent and which decision step led to the failure? Existing methods suffer from severe limitations. For example, in a failed web search task, existing methods often attribute the failure to the final agent that reports ``no results found," when the actual root cause may be an upstream agent that formulated an overly restrictive search query. Manual analysis by domain experts is not only labor-intensive and costly, but also difficult to scale. Statistical correlation-based diagnostic methods fail to effectively detect true root causes, often misidentifying downstream "symptoms" (such as agents producing final erroneous outputs) as upstream "root causes," thereby generating misleading attribution results~\cite{jiang2023cf,epperson2025interactive,jiang2025codejudgebench}. Although recent ``LLM-as-Judge''(where large language models are employed to analyze execution traces and directly judge which agents or steps are responsible for failures) approaches achieve automation, their accuracy remains concerning. For instance, on the specialized Who\&When benchmark \cite{zhang2025agent}, existing best methods achieve only \textbf{17.1\%} accuracy in locating critical error steps, while on the more complex TRAIL benchmark \cite{deshpande2025trail}, top-performing models achieve a joint accuracy as low as \textbf{18.3\%}. This substantial ``attribution gap" indicates that we need a fundamentally new, more principled approach to address this problem. 

As shown in figure \ref{fig1} To bridge this gap, we advocate for a paradigm shift from ``correlation'' analysis to ``causal'' reasoning.Many typical failure modes in MAS involve complex causal chains—for instance, when an upstream data-collection agent provides incomplete information, causing a downstream analysis agent to make incorrect decisions, which then propagates to subsequent agents and ultimately leads to task failure. These cascading dependencies form intricate causal networks that cannot be resolved by analyzing surface-level data flows or activity patterns in logs. Unlike traditional methods that merely capture surface associations, causal inference provides a rigorous theoretical toolkit capable of handling confounding variables, modeling dynamic dependencies, and identifying true root causes from complex interaction networks~\cite{pearl2009causality,witmer2020book}. Many typical failure modes in MAS, such as downstream agent decision failures caused by upstream agents information withholding or provision of erroneous data~\cite{park2023generative}, are essentially causal propagation problems. These issues cannot be resolved by analyzing data flows or surface activities in logs. Therefore, a causal framework that correctly models performance impacts rather than data movements is crucial for accurately diagnosing these cascading and synergistic failures. \cite{chen2025understanding}

This paper presents the first multi-granularity causal inference framework for automated failure attribution in MAS, aimed at providing precise, interpretable, and actionable diagnostics. Our approach encompasses innovations at two levels. At the agent level, we design the \textbf{SBSLocator} module, which introduces the ``Performance Causal Inversion'' principle, constructing correct performance causal graphs by inverting observed data dependencies to trace back to failure origins. Additionally, it incorporates \textbf{Shapley values} to quantify the systematic importance of each agent in complex collaborations. At the step level, the \textbf{CPIdentifier} module achieves precise prediction of critical error steps by constructing a multi-dimensional feature space encompassing technical complexity, interaction dynamics, and semantic consistency, and utilizing our proposed \textbf{CDC-MAS algorithm} to discover temporal causal structures and handle confounding factors in non-stationary environments. The entire framework is centered on \textbf{counterfactual reasoning}, quantifying the true impact of agents/steps on task failure by simulating scenarios of ``what would happen if a certain agent/step performed normally?''


The main contributions of this paper are as follows:

\begin{itemize}
    \item \textbf{Causal Attribution Framework}: First systematic application of causal inference and counterfactual reasoning to MAS failure attribution, overcoming the limitations of traditional correlation-based methods.
    
    \item \textbf{Multi-Granularity Causal Modeling}: Design of a hierarchical causal discovery pipeline that simultaneously captures agent-level bottlenecks and step-level critical decisions, enabling fine-grained diagnosis across temporal, spatial, and functional scales.
    
    \item \textbf{Performance Causal Inversion and Shapley Value Integration}: Proposal of a novel attribution mechanism that combines the performance causal inversion principle with Shapley value analysis to accurately distinguish root causes from surface symptoms and quantify agents' true influence in collaboration.
    
    \item \textbf{Validation of Actionable Diagnostics}: Experimental demonstration that optimization suggestions guided by our causal attribution can significantly improve task success rates, showcasing the tremendous potential of this framework in practical debugging and optimization workflows.
\end{itemize}
\begin{figure*}[t]
\centering
\includegraphics[width=\linewidth]{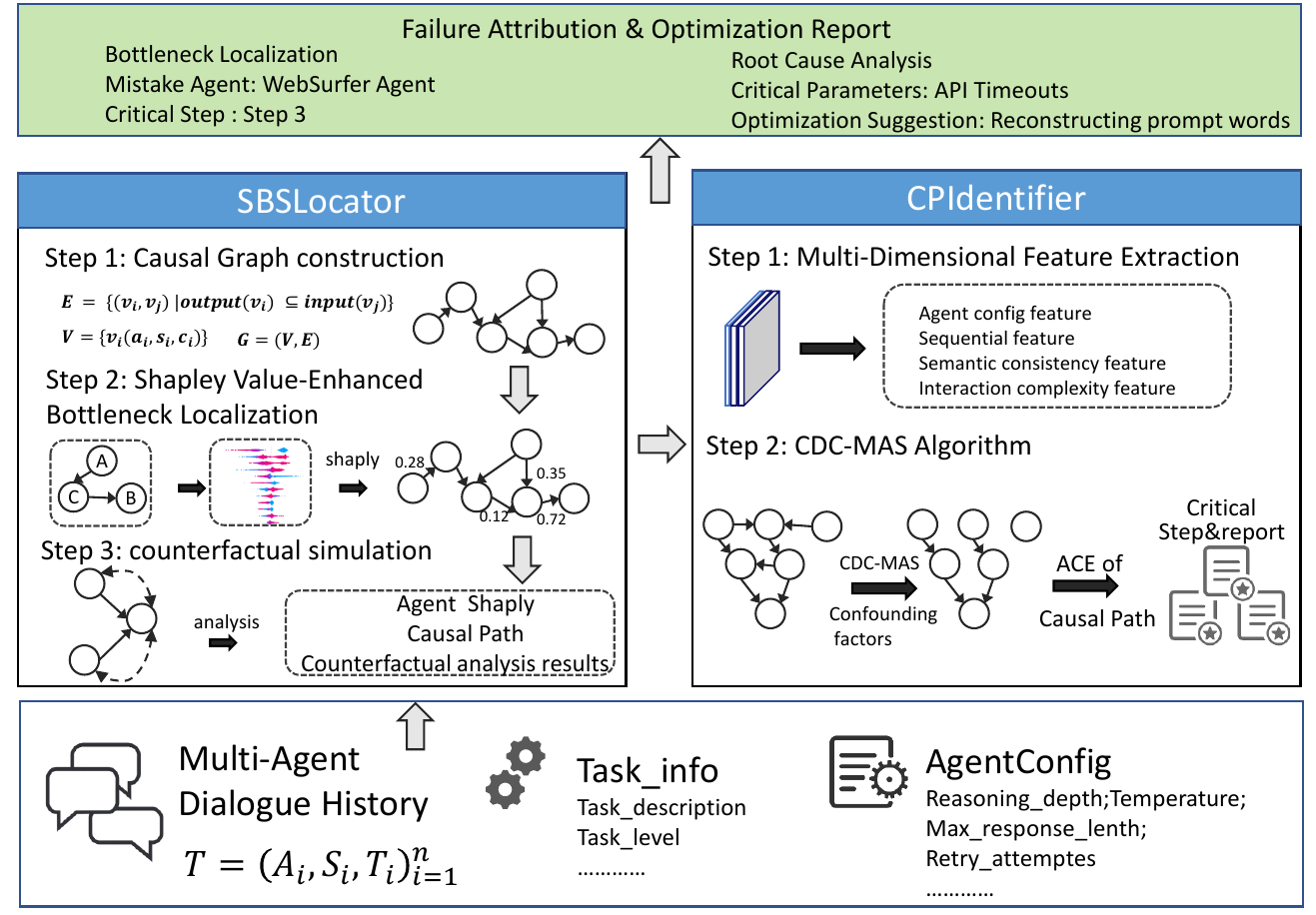} 
\caption{Failure attribution process based on performance causality inversion and hierarchical causal discovery. This figure illustrates how our method transforms the original execution trace into an actionable diagnostic report through a two-stage pipeline. Key innovations include the SBSLocator module for agent attribution and the CPIdentifier module for identifying critical steps, with the latter integrating our CDC-MAS algorithm to handle confounding factors in non-stationary data.}
\label{fig2}
\end{figure*}
\section{Methodology}

\subsection{Problem Formalization}

The execution process of a multi-agent system (MAS) is inherently a complex temporal causal system, where interactions and decisions among agents form an intricate dependency network. Given a failed execution trajectory $T = \langle(a_i, s_i, t_i, c_i)\rangle_{i=1}^{n}$, where $a_i \in \mathcal{A}$ represents the agent at the $i$-th step, $s_i \in \mathcal{S}$ represents the specific decision action, $t_i \in \mathbb{R}^+$ represents the execution time, and $c_i \in \mathcal{C}$ represents the system context state at that time, the core challenge we face is to accurately identify the key agent responsible for the task failure (Who problem) and the decisive error step (When problem).

However, the confounding phenomena prevalent in MAS environments make this task extremely challenging. These confounding factors can be divided into two categories: one is observable confounding factors, such as task complexity $\tau \in \mathbb{R}^+$, which simultaneously affects the decision difficulty of multiple agents and the final success probability, as well as the agent configuration parameter vector $\theta = \{\theta_{\text{reasoning}}, \theta_{\text{temperature}}, \theta_{\text{timeout}}, \dots \}$, which has a systemic impact on the entire execution process; the other is latent confounding factors $\mathcal{U}$, such as performance fluctuations of external APIs or changes in emergent collaboration patterns among agents. Although these factors are not directly observed or recorded, they leave hidden causal traces in the execution trajectory. It is precisely these confounding factors that cause traditional statistical correlation-based analysis methods to produce a large number of false associations, thereby misleading the accuracy of failure attribution. As shown in Figure 2, our framework addresses these challenges through a two-stage causal inference pipeline.

\subsection{SBSLocator: Agent-Level Attribution Based on Performance Causal Inversion}

To address the "Who" problem, we designed the SBSLocator module, which identifies the bottleneck agent causing system failure through a novel causal principle, combined with game-theoretic quantification methods and counterfactual reasoning.

\subsubsection{Theoretical Foundation of Performance Causal Inversion }

Before delving into the technical details, we must first elucidate a counterintuitive but crucial phenomenon in MAS: the direction of information flow is opposite to the direction of performance causality. Its root lies in the dependency propagation mechanism among agents. Consider a typical scenario: upstream agent A (e.g., data collection agent) transmits information to downstream agent B (e.g., data analysis agent). From the data flow perspective, information flows from A to B; however, from the performance causality perspective, if A provides erroneous or incomplete data, leading to B's subsequent analysis failure, then B's performance manifestation is actually causally determined by A's performance deficiency. In other words, A's performance state is the root cause of B's performance manifestation.

Based on this insight, we propose the \textbf{Performance Causal Inversion (PCI)} principle: to construct an accurate performance causal graph, the data dependencies observed from the execution logs must be reversed in direction.

\noindent\textbf{Definition 1 (Data Dependency Graph and Performance Causal Graph):} Given the execution trajectory $T$, we first construct a data dependency graph $G_{\text{data}} = (V, E_{\text{data}})$: the node set $V = \{v_i = (a_i, s_i, c_i)\}$ represents agent-step-context triples. The edge set $E_{\text{data}} = \{(v_i, v_j) \mid \text{output}(v_i) \subseteq \text{input}(v_j) \}$ represents information transmission relations. Then, through causal inversion transformation, we obtain the performance causal graph: $G_{\text{causal}} = (V, E_{\text{causal}})$; where $E_{\text{causal}} = \{(v_j, v_i) \mid (v_i, v_j) \in E_{\text{data}}\}$. This transformation ensures that each edge in the causal graph points from the performance cause to the performance effect, establishing a correct structural foundation for subsequent causal inference.

Although the PCI principle is intuitive and powerful, its application relies on key assumptions such as the acyclicity of the performance graph, and has limitations in scenarios with complex feedback loops or collaborative failures. For detailed discussion of the theoretical boundaries and potential failure modes of this principle, refer to Appendix A.

\subsubsection{System Importance Quantification Combined with Shapley Values}

Merely identifying causal relationships is insufficient; we also need to quantify each agent's marginal contribution to the entire system. To this end, we introduce Shapley values from game theory, treating each agent as a participant in a "cooperative game" and evaluating its true system importance by calculating its marginal contributions in different coalition combinations.

\textbf{Definition 2 (Shapley-Enhanced Functional Causal Model):} For each node \(j\) in the performance causal graph, its state is generated by the following enhanced structural equation:
\begin{equation}
    X_j := f_j(\mathrm{PA}_j, N_j) \cdot \exp(\alpha \cdot  \hat{\phi}_j) 
\end{equation}

The design logic of this model is that the final performance \(X_j\) of agent \(j\) is determined by three factors: \(f_j(\mathrm{PA}_j, N_j)\): base performance, depending on the input quality from parent nodes (\(\mathrm{PA}_j\)) and the agent's own processing capability (\(N_j\)). \(\exp(\alpha \cdot \hat{\phi}_j)\): system importance amplification factor, where the higher the Shapley value \(\hat{\phi}_j\) of the agent, the more significant its performance fluctuations on the system.

Due to the exponential complexity of exact Shapley value computation, we employ Monte Carlo sampling for approximation, reducing the complexity to polynomial level.

\subsubsection{Bottleneck Identification Based on Counterfactual Reasoning}

With the causal graph structure and system importance quantification, we employ counterfactual reasoning to determine which agent is the true ``culprit" of the task failure. We pose a key question for each suspect agent: ``If this agent performs normally, would the entire task succeed?"

\textbf{Definition 3 (Integrated Bottleneck Score):} We calculate the bottleneck score \(\mathrm{BS}_j\) for each agent \(j\) by integrating system importance and counterfactual improvement potential through the following formula:
\begin{equation}
    \mathrm{BS}_j = \hat{\phi}_j \times (Y_j^{\text{cf}} - Y^{\text{original}}) \times I 
\end{equation}

where: \(\hat{\phi}_j\) is the Shapley value of agent \(j\). \(Y^{\text{original}}\) is the actual task outcome observed in the failed trajectory. \(Y_j^{\text{cf}}\) is the counterfactual outcome, simulating the task result when agent \(j\) performs normally. \(\theta_{\text{success}}\) is a predefined success threshold. \(I[\cdot]\) is the indicator function. The design philosophy of this score is: only agents that have both high system importance and can significantly improve the task outcome through self-correction are identified as true bottlenecks.

\subsection{CPIdentifier: Critical Step Prediction Based on Causal Deconfounding}

After identifying the bottleneck agent, the CPIdentifier module focuses on precisely locating the key error step in that agents execution trajectory . This process is achieved through a multi-stage causal discovery algorithm CDC-MAS, specifically designed to handle the non-stationarity and confounding biases prevalent in MAS interaction data.

\subsubsection{Multi-Dimensional Feature Space Construction}

To comprehensively capture the contextual information of each step, CPIdentifier first extracts multi-dimensional features from the failed trajectory \(T\). We model the importance of steps as a function of four dimensions:
\begin{equation}
    F = F_{\text{tech}} \oplus F_{\text{interact}} \oplus F_{\text{temporal}} \oplus F_{\text{semantic}} 
\end{equation}

Technical Complexity (\(F_{\text{tech}}\)): Quantifies the intrinsic computational difficulty of the step. Interaction Complexity (\(F_{\text{interact}}\)): Captures agent collaboration effects. Temporal Dynamics (\(F_{\text{temporal}}\)): Reflects execution rhythm. Semantic Consistency (\(F_{\text{semantic}}\)): Evaluates the semantic quality of output content. These features provide rich, context-aware inputs for subsequent causal discovery.

\subsubsection{CDC-MAS Algorithm: A Phased Causal Discovery Process}

The CDC-MAS algorithm is the core of our technical contribution, robustly discovering causal structures and predicting key steps from high-dimensional features through a clear four-stage process. For its pseudocode, please refer to Appendix B.

High-Level Overview and Process: The algorithm's process can be summarized as: (I) Feature Engineering → (II) Temporal Skeleton Discovery → (III) Confounding-Aware Edge Orientation → (IV) Causal Effect Ranking. Each stage builds on the output of the previous one, progressively refining to the final key step ranking.

Stage I: Context-Aware Feature Preparation. \textbf{Objective}: Prepare clean, information-rich feature data and explicitly encode confounding factors affecting the entire trajectory. \textbf{Method:} This stage extracts multi-dimensional features and constructs a learnable context vector \(C_{\text{MAS}}\) to capture global confounding information. This vector is generated by a Transformer encoder:
\begin{equation}
    C_{\text{MAS}} = \text{Transformer}([F_{\text{task}}; F_{\text{config}}; F_{\text{dynamic}}]) 
\end{equation}

Stage II: Temporal Causal Structure Discovery. \textbf{Objective:} Build an initial graph skeleton containing all potential temporal dependencies, while handling non-stationarity by conditioning on \(C_{\text{MAS}}\). \textbf{Method:} This stage constructs an initial graph skeleton containing potential temporal dependencies through a series of conditional independence tests conditioned on \(C_{\text{MAS}}\).

Stage III: Confounding-Aware Edge Orientation. \textbf{Objective:} Orient the undirected edges left from the previous stage using \(C_{\text{MAS}}\) as a proxy for confounding information, thereby distinguishing causal relations from spurious associations. \textbf{Method:} Utilize the context vector \(C_{\text{MAS}}\) as a proxy for confounding information to orient undirected edges, distinguishing causal relations from spurious associations.

Stage IV: Causal Path Analysis and Key Step Ranking. \textbf{Objective:} On the final determined causal graph, quantify the average causal effect (ACE) of each step on the task failure and rank accordingly. \textbf{Method:} For each step \(i\), calculate its ACE score. This score measures the comprehensive impact of all causal paths starting from this step on the final outcome:
\begin{equation}
    \text{ACE}_i = \frac{\sum_{\text{path} \in P_i^{\text{from } G_{\text{causal}}}} \prod_{e \in \text{path}} \text{LocalEffect}(e)}{\text{Context\_Weight}(e, C_{\text{MAS}})}
    \end{equation}
To ensure reproducibility, we clearly define: \(\text{LocalEffect}(e)\) is estimated by the coefficient in a local linear regression model; \(\text{Context\_Weight}(e, C_{\text{MAS}})\) is a context modulation function based on cosine similarity.

\textbf{Theorem 1 (Theoretical Guarantee):} Under the assumptions of causal Markov property, faithfulness, and contextual sufficiency, CDC-MAS can asymptotically consistently recover the true causal structure (Proof and related analysis can be found in Appendix B).

\subsubsection{Counterfactual Validation and Robustness Guarantees}

To verify the reliability of the initial ranking, CPIdentifier introduces counterfactual interventions to further optimize the predictions. Counterfactual Step Intervention: For each candidate step \(s_k\), we simulate an ``optimal" intervention and calculate its expected improvement on the final outcome \(\Delta_k\):
\begin{equation}
    \Delta_k = \mathbb{E}[Y | \text{do}(X_k = x_k^{\text{optimal}})] - \mathbb{E}[Y | X_k = x_k^{\text{observed}}]
\end{equation}

Final Ranking: The final score integrates causal effect, improvement potential, and uncertainty:
\begin{equation}
    \text{FinalScore}_k = \omega_1 \cdot \text{ACE}_k + \omega_2 \cdot \Delta_k + \omega_3 \cdot \text{Confidence}_k \quad 
\end{equation}

where \(\text{Confidence}_k\) is the prediction confidence estimated via bootstrap resampling. This final ranking is not only causally consistent but also has practical improvement potential, providing reliable guidance with confidence levels for subsequent MAS optimization.

\section{Experiments}

This section evaluates the proposed framework's fault attribution performance on multi-agent system (MAS) failure trajectories through quantitative and qualitative analysis. We particularly emphasize the contributions of our core methodological innovations, such as performance causal inversion, Shapley value quantification, and counterfactual reasoning, in handling cascading failures and confounding factors. We validate the framework's accuracy, actionability, and component contributions via benchmark comparisons, optimization experiments, and ablation studies.

\subsection{Experimental Setup}

\subsubsection{Datasets}
To ensure a comprehensive and diverse evaluation, we selected two complementary benchmark datasets that cover varying task complexities, trajectory lengths, and interaction types:
\begin{itemize}
    \item \textbf{Who\&When} \cite{zhang2025agent}: This dataset contains 184 failure instances from 127 LLM-based multi-agent systems. It is divided into two subsets: Algo-Generated systems (5-10 steps), which focus on short-chain failures, and Hand-Crafted systems (5-130 steps), which feature long-range dependencies and complex interactions. The tasks include open-world question answering  and code debugging. The dataset provides human-annotated ground truth for the responsible agent and the critical error step, making it ideal for validating our framework's ability to attribute ``Who" (agent-level) and ``When" (step-level) in cascading failures. We pay special attention to the challenging Hand-Crafted subset to test the framework's performance in realistic scenarios.
    \item \textbf{TRAIL} \cite{deshpande2025trail}: A benchmark containing 118 traces from GAIA-based open-world information retrieval tasks. With an average trajectory length of 8.28 spans, this dataset emphasizes error localization in dynamic, multi-modal, and interactive workflows. It further validates the framework's step-level precision in environments with unstructured data and real-time decisions.
\end{itemize}
This selection of datasets ensures a progressive evaluation from simple to complex scenarios and allows us to analyze the framework's sensitivity to trajectory length and interaction depth. For Who\&When, we adopt an 80/10/10 split for training, validation, and testing.

\subsubsection{Baseline Methods}
We compare our framework against several baselines, including a random baseline and state-of-the-art ``LLM-as-Judge" methods, all implemented using GPT-4o to highlight the advantages of our causal paradigm:
\begin{itemize}
    \item \textbf{Random}: Randomly selects an agent or step as the source of failure, serving as a lower-bound baseline to quantify the effectiveness of non-causal approaches.
    \item \textbf{All-at-Once}: Analyzes the entire trajectory at once to make an attribution, representing a global correlation-based method.
    \item \textbf{Step-by-Step}: Sequentially examines each step, simulating a gradual diagnostic process.
    \item \textbf{Binary Search}: Uses binary search to efficiently locate the error, optimizing search efficiency but still relying on surface-level correlations.
\end{itemize}
These baselines highlight the limitations of traditional correlation-based analysis, such as ignoring confounding factors and causal chains. All methods were run on identical hardware and with the same parameter settings to ensure a fair comparison.

\subsubsection{Evaluation Metrics}
We adopt the standard metrics from the Who\&When benchmark and extend our evaluation to include statistical significance testing (t-test, $p<0.05$):
\begin{itemize}
    \item \textbf{Agent-Level Accuracy}: The proportion of instances where the failure-responsible agent is correctly identified. This evaluates the precision of "Who" attribution.
    \item \textbf{Step-Level Accuracy}: The proportion of instances where the decisive error step is correctly located. This evaluates the precision of "When" attribution.
\end{itemize}
All reported results are the mean $\pm$ standard deviation of 5 independent runs to enhance reliability.

\subsection{Main Results: Attribution Accuracy}
We evaluated the performance of our full framework against the baselines on both datasets. The results for the Who\&When benchmark, with a clear distinction between the Algo-Generated and Hand-Crafted subsets, are presented in Table~\ref{tab:who_when_comparison}.

\begin{table}[h!]
\centering
\caption{Performance comparison on the Who\&When benchmark (Algo/Hand subsets). Best performance is in \textbf{bold}.}
\label{tab:who_when_comparison}
\resizebox{\columnwidth}{!}{
\begin{tabular}{lcc}
\hline
\textbf{Method} & \textbf{Agent Acc. (\%)} & \textbf{Step Acc. (\%)} \\
 & \textbf{(Algo/Hand)} & \textbf{(Algo/Hand)} \\ \hline
Random & 29.10 / 12.00 & 19.06 / 4.16 \\
All-at-Once & \textbf{54.33} / 55.17 & 12.50 / 5.26 \\
Step-by-Step & 35.20 / 34.48 & 25.51 / 7.02 \\
Binary Search & 44.13 / 51.72 & 23.98 / 6.90 \\
\textbf{Ours} & 48.5 / \textbf{56.8} & \textbf{36.2 / 18.2} \\ \hline
\end{tabular}}
\end{table}

As shown in Table~\ref{tab:who_when_comparison}, our framework consistently outperforms all baselines across all metrics and subsets. The improvement is particularly pronounced on the more challenging Hand-Crafted subset (long trajectories), where our Step-Level Accuracy reaches \textbf{18.2\%}, achieving a relative improvement of over 159\% compared to the best baseline (7.02\% from Step-by-Step). This demonstrates the superiority of causal reasoning in distinguishing root causes from superficial correlations, especially when dealing with complex, long-range dependencies. Qualitative analysis reveals that baselines often misattribute the failure to downstream ``symptoms" (e.g., the final agent that produced an error), whereas our method, through performance causal inversion, successfully traces the issue back to its upstream ``pathogen." All key improvements are statistically significant at the $p<0.05$ level.

\begin{figure}[t]
\centering
\includegraphics[width=1\columnwidth]{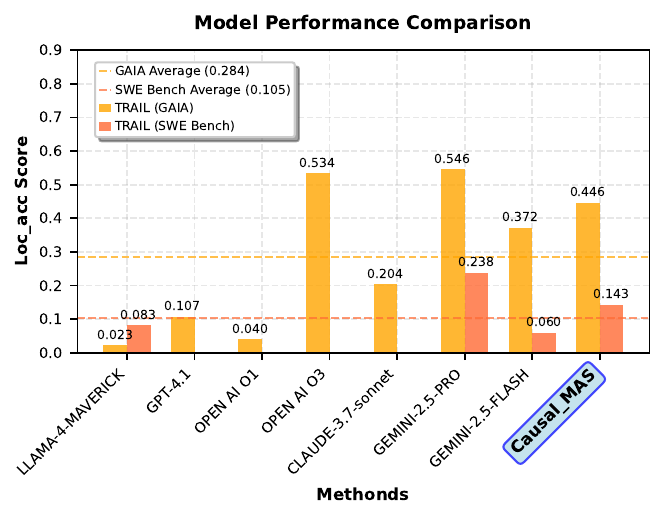} 
\caption{ Step-level attribution accuracy on the TRAIL benchmark.}
\label{trail_performance}
\end{figure}
On the TRAIL benchmark, our framework achieves a step-level accuracy of 44.6\% on GAIA and 14.3\% on SWE Bench, substantially outperforming the average baseline performances of 24.8\% and 10.5\% respectively. As illustrated in Figure \ref{trail_performance}, the Causal-MAS bars clearly exceed those of all competing models across both subsets, highlighting the effectiveness of our causal reasoning framework in identifying failure-critical steps in complex multi-agent workflows.

\subsection{Operability: Closed-Loop Optimization}

To move beyond accuracy metrics and assess the practical utility of our framework, we designed and executed a closed-loop optimization experiment. This test verifies whether our causal diagnoses can effectively guide automated repairs to improve final task success rates, demonstrating a shift from passive analysis to active problem-solving.

The experimental process for our Causal Repair method unfolds in three stages:
\begin{itemize}
  \item \textbf{Diagnosis:} We selected 50 failed trajectories from the \textit{Who\&When} dataset. For each failure, our framework generated a precise root cause report, such as:  
    ``The WebSurfer agent failed at step 3 because its API call instruction was ambiguous.”

  \item \textbf{Repair:} An optimization agent, powered by GPT-4o, then used this targeted diagnosis to automatically generate a revised solution. For instance, following the diagnosis above, it would refine the vague instruction “search for Oscar-winning films” into the specific “search for 2023 Oscar-winning films,” directly addressing the identified problem.
  \item \textbf{Validation:} Finally, the multi-agent task was re-executed from the newly repaired step, and the new overall task success rate was measured to quantify the repair’s effectiveness.
\end{itemize}

To rigorously evaluate the impact of our Causal Repair, we compared its performance against two control conditions: the original, uncorrected success rate (Base SR) and a Random Repair (Rand.\ SR), where a random, non-targeted modification is applied to simulate a blind, trial-and-error debugging attempt.

\begin{table}[h!]
  \centering
  \caption{Success rates (SR) after closed-loop optimization. All values are percentages (\%).}
  \label{tab:compact_optimization}
  \small
  \resizebox{\columnwidth}{!}{
  \begin{tabular}{@{}lccc@{}}
    \toprule
    \textbf{Task Type}    & \textbf{Base SR} & \textbf{Rand.\ SR} & \textbf{Causal SR} \\
    \midrule
    Prompt Fails          & 12.5             & 15.2               & \textbf{35.4}      \\
    Tool Fails            & 18.2             & 21.3               & \textbf{40.2}      \\
    \midrule
    \textbf{Overall}      & \textbf{15.4}    & \textbf{18.3}      & \textbf{37.8}      \\
    \bottomrule
  \end{tabular}}
\end{table}

The results in Table~\ref{tab:compact_optimization} confirm our method’s significant practical value. Repairs guided by our causal analysis boosted the overall task success rate from 15.4\% to 37.8\%—an absolute improvement of 22.4 percentage points.

This result far surpasses the Random Repair control, which only reached an 18.3\% success rate. The stark contrast proves that the performance gain stems from our framework’s accurate, targeted guidance, not merely the act of intervention. This experiment successfully closes the ``diagnose–validate–optimize'' loop and validates our method’s tangible benefits for automated debugging.

\subsection{Ablation Study}
We conducted an ablation study to isolate and verify the contributions of each key component in our framework. We systematically removed one module at a time and evaluated the performance degradation on the Algo-Generated subset of the Who\&When dataset. The results are presented in Table~\ref{tab:ablation_study}.

\begin{table}[h!]
\centering
\caption{Ablation study results on the Who\&When (Algo-Generated) dataset. Performance drops confirm the necessity of each component.}
\label{tab:ablation_study}
\resizebox{\columnwidth}{!}{
\begin{tabular}{lcc}
\hline
\textbf{Variant} & \textbf{Agent Acc. } & \textbf{Step Acc. } \\
 & \textbf{(Drop)} & \textbf{(Drop)} \\ \hline
\textbf{Full Model} & \textbf{48.5} & \textbf{36.2} \\
w/o Causal Inversion & 40.7 (-7.8) & 28.1 (-8.1) \\
w/o Shapley Value & 42.1 (-6.4) & 24.5 (-11.7) \\
w/o Context Conditioning & 40.7 (-7.8) & 22.3 (-13.9) \\
w/o Counterfactual Reasoning & 45.6 (-2.9) & 30.4 (-5.8) \\ \hline
\end{tabular}}
\end{table}

As Table~\ref{tab:ablation_study} shows, every component is critical to the framework's overall performance:
\begin{itemize}
    \item \textbf{Removing Causal Inversion} causes a significant drop in Agent-Level Accuracy (-7.8 pts), confirming its core role in reversing data flow dependencies to correctly trace performance causality and avoid misjudging downstream symptoms as root causes.
    \item \textbf{Removing Shapley Values} leads to the second-largest drop in Step-Level Accuracy (-11.7 pts), highlighting the necessity of quantifying marginal contributions to assess each agent's true influence in complex collaborations.
    \item \textbf{Removing Context Conditioning} results in the most severe drop in Step-Level Accuracy (-13.9 pts), validating its critical contribution to handling non-stationary data and confounding factors in complex systems.
    \item \textbf{Removing Counterfactual Reasoning} also leads to a notable performance decline, demonstrating the practical utility of ``what-if" simulations for validating and refining attribution rankings.
\end{itemize}
These results clearly demonstrate that the framework's components work synergistically, and their integration is essential for achieving accurate and robust fault attribution.

\section{Related Work}

\subsection{Multi-Agent Systems}
Large Language Models (LLMs) have been increasingly used as central controllers for developing agents that interact with the external world beyond text \cite{achiam2023gpt, wang2024comprehensive, deng2023mind2web, xie2024osworld, zhang2025nemotron, zhang2024offline}. While single-agent systems excel in specific tasks \cite{yao2023react, zhang2023ecoassistant, zhang2024ecoact}, they falter in scenarios requiring collaboration. To overcome this, LLM-powered multi-agent systems (MAS) enable concurrent interactions among specialized agents , simulating real-world cooperation for complex problem-solving. Frameworks like CAMEL \cite{li2023camel}, ChatDev \cite{qian2023chatdev}, and MetaGPT \cite{hong2023metagpt} assign roles and incorporate Standard Operating Procedures (SOPs) for structured coordination in areas like software development, while Generative Agents model emergent dynamics via memory and reflection. However, MAS exhibit high failure rates (40-60\%), with taxonomies like MAST classifying issues into specification/design failures, inter-agent misalignments, and verification errors. Benchmarks such as AgentBench \cite{liu2023agentbench} assess performance but lack causal diagnostic tools, emphasizing the need for automated failure attribution in interaction-heavy systems.

\subsection{LLM-as-Judge for Failure Attribution}
LLMs have been widely adopted as evaluators for tasks based on predefined standards \cite{fu2023gptscore, gu2024survey, hu2024language, li2023alpacaeval, liu2023g, thakur2024judging}, reducing human labor in areas like chat evaluation \cite{chan2023chateval, zheng2023judging} and text summarization \cite{miao2023selfcheck, van2024field}. In agentic systems, LLMs-as-judges analyze feedback for corrections \cite{shinn2023reflexion} and evaluate workflows in datasets like DevAI \cite{zhuge2024agent}. For failure attribution in MAS, approaches on benchmarks like Who\&When and TRAIL use strategies such as All-at-Once, Step-by-Step, and Binary Search, but achieve low accuracies due to correlation-based limitations, often confusing symptoms with causes. Our work advances this by integrating Shapley values from game theory for fair responsibility allocation and causal inference with performance causal inversion, enabling robust root-cause identification in non-stationary MAS interactions.

\section{Conclusion}


Existing MAS failure attribution methods suffer from critical limitations: manual analysis is unscalable, while correlation-based approaches achieve less than 20\% accuracy by confusing symptoms with root causes. LLM-as-Judge methods remain inadequate due to their inability to model causal dependencies.
We present the first multi-granularity causal inference framework for MAS failure attribution. Key innovations include performance causal inversion for accurate causal graphs, Shapley values for agent contribution quantification, and the CDC-MAS algorithm for non-stationary data. Evaluations show up to 36.2\% step-level accuracy, significantly outperforming baselines. Our closed-loop experiments demonstrate actionability, with targeted optimizations boosting task success rates by 22.4\%.
This work establishes causal reasoning as a principled pathway to reliable multi-agent systems. Future directions include extending to multi-modal interactions, developing online causal discovery for real-time attribution, and integrating with automated self-healing mechanisms.

\bibliography{AnonymousSubmission/LaTeX/reference}

\newpage
\clearpage

\newpage

\setlength{\leftmargini}{20pt}
\makeatletter\def\@listi{\leftmargin\leftmargini \topsep .5em \parsep .5em \itemsep .5em}
\def\@listii{\leftmargin\leftmarginii \labelwidth\leftmarginii \advance\labelwidth-\labelsep \topsep .4em \parsep .4em \itemsep .4em}
\def\@listiii{\leftmargin\leftmarginiii \labelwidth\leftmarginiii \advance\labelwidth-\labelsep \topsep .4em \parsep .4em \itemsep .4em}\makeatother

\setcounter{secnumdepth}{0}
\renewcommand\thesubsection{\arabic{subsection}}
\renewcommand\labelenumi{\thesubsection.\arabic{enumi}}

\newcounter{checksubsection}
\newcounter{checkitem}[checksubsection]

\newcommand{\checksubsection}[1]{%
  \refstepcounter{checksubsection}%
  \paragraph{\arabic{checksubsection}. #1}%
  \setcounter{checkitem}{0}%
}

\newcommand{\checkitem}{%
  \refstepcounter{checkitem}%
  \item[\arabic{checksubsection}.\arabic{checkitem}.]%
}
\newcommand{\question}[2]{\normalcolor\checkitem #1 #2 \color{blue}}
\newcommand{\ifyespoints}[1]{\makebox[0pt][l]{\hspace{-15pt}\normalcolor #1}}

\end{document}